\documentclass[letterpaper, 10 pt, conference]{ieeeconf}
\usepackage[utf8]{inputenc}
\usepackage[T1]{fontenc}
\IEEEoverridecommandlockouts                              
\usepackage{cite}
\usepackage{amsmath,amssymb,amsfonts}
\usepackage{algorithmic}
\usepackage{graphicx}
\usepackage{placeins}
\usepackage{booktabs}
\usepackage{multirow}
\usepackage{textcomp}
\usepackage{xcolor}
\def\BibTeX{{\rm B\kern-.05em{\sc i\kern-.025em b}\kern-.08em
T\kern-.1667em\lower.7ex\hbox{E}\kern-.125emX}}
\newcommand{\naiveasync}{Na\"{\i}ve Async }
\begin{document}

\title{\LARGE \bf
Action ControlNet: A Lightweight Delay-Aware Adapter for Smooth Asynchronous Control in
Vision-Language-Action Models }

\author{Tiecheng Guo and Meng Guo$^{*}$%
\thanks{The authors are with the School of Advanced Manufacturing and Robotics,
  Peking University, Beijing 100871, China.}%
\thanks{$^{*}$Corresponding author: Meng Guo, \texttt{meng.guo@pku.edu.cn}.}
}

\maketitle
\thispagestyle{empty}
\pagestyle{empty}

\begin{abstract}
Vision-language-action (VLA) models have shown strong potential for general-purpose
robot manipulation, but their inference latency remains a major obstacle to stable
high-frequency control. Asynchronous execution mitigates this bottleneck by overlapping
policy inference with action execution, yet the next action chunk is still predicted
from stale observations while the robot continues to move. Direct chunk stitching
therefore introduces handoff discontinuities, action jitter, and failures in
contact-rich manipulation. Existing remedies typically require either full-policy
retraining or architecture-specific runtime logic. This work proposes Action ControlNet
(ACNet), a lightweight delay-aware adapter that uses the executed motion suffix as a
residual condition for a mostly frozen action head. ACNet leaves the pretrained backbone
unchanged, introduces few trainable parameters, and remains compatible with generative
action heads such as diffusion and flow matching. On Kinetix, Meta-World MT50, and a
real-world SO-ARM101 platform, ACNet improves robustness under inference delay and
yields smoother asynchronous trajectories than direct chunk stitching, while remaining
more lightweight than full delay-conditioned retraining.
\end{abstract}


\section{Introduction}

Vision-language-action (VLA) models~\cite{b1} have recently emerged as a promising
framework for general-purpose robot manipulation, benefiting from large-scale multimodal
pretraining that supports broad task coverage and instruction following across diverse
scenarios. Modern VLA policies often pair large vision-language backbones with
generative action heads, such as diffusion~\cite{b2} or flow matching~\cite{b3}, and
generate short action chunks rather than individual actions. In the standard synchronous
deployment loop, the robot first waits for policy inference to finish and only then
executes the predicted chunk. Because inference latency is non-negligible for large
backbones and iterative action heads, this infer-then-execute schedule introduces idle
intervals between consecutive chunks. In practice, these idle intervals appear as
visible stop-and-go pauses during manipulation, making the robot motion less continuous
and increasing the wall-clock time required to complete a task. This synchronous latency
bottleneck motivates asynchronous execution, where model inference is overlapped with
robot motion.

\begin{figure}[!t]
\centering
\includegraphics[width=0.98\columnwidth]{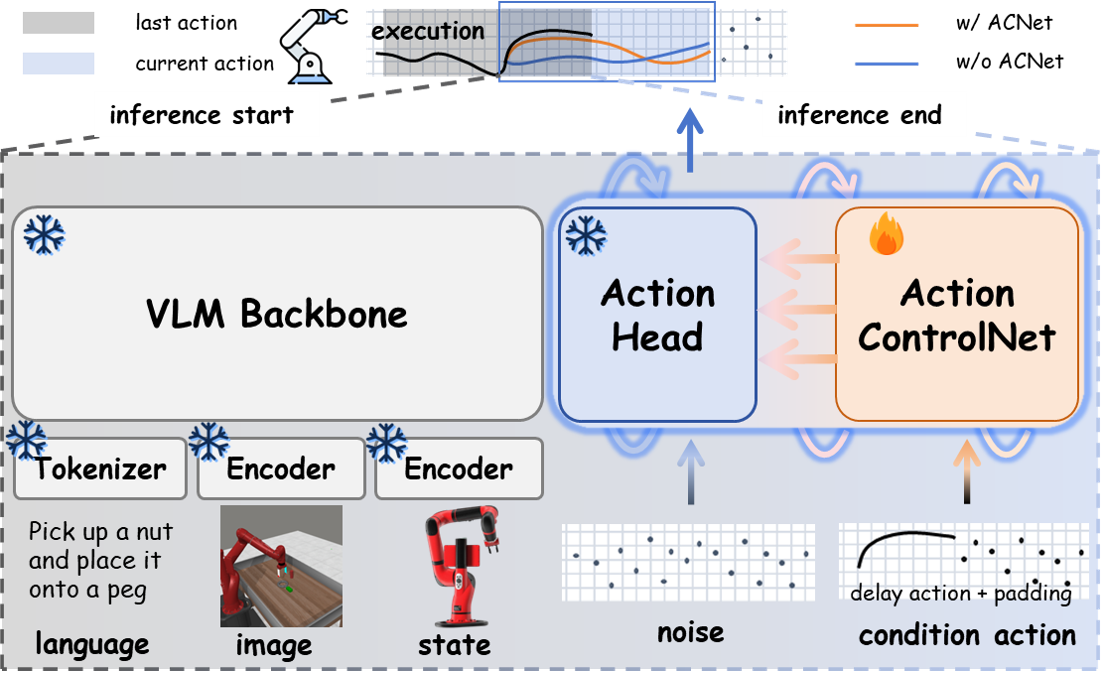}

\includegraphics[width=0.98\columnwidth]{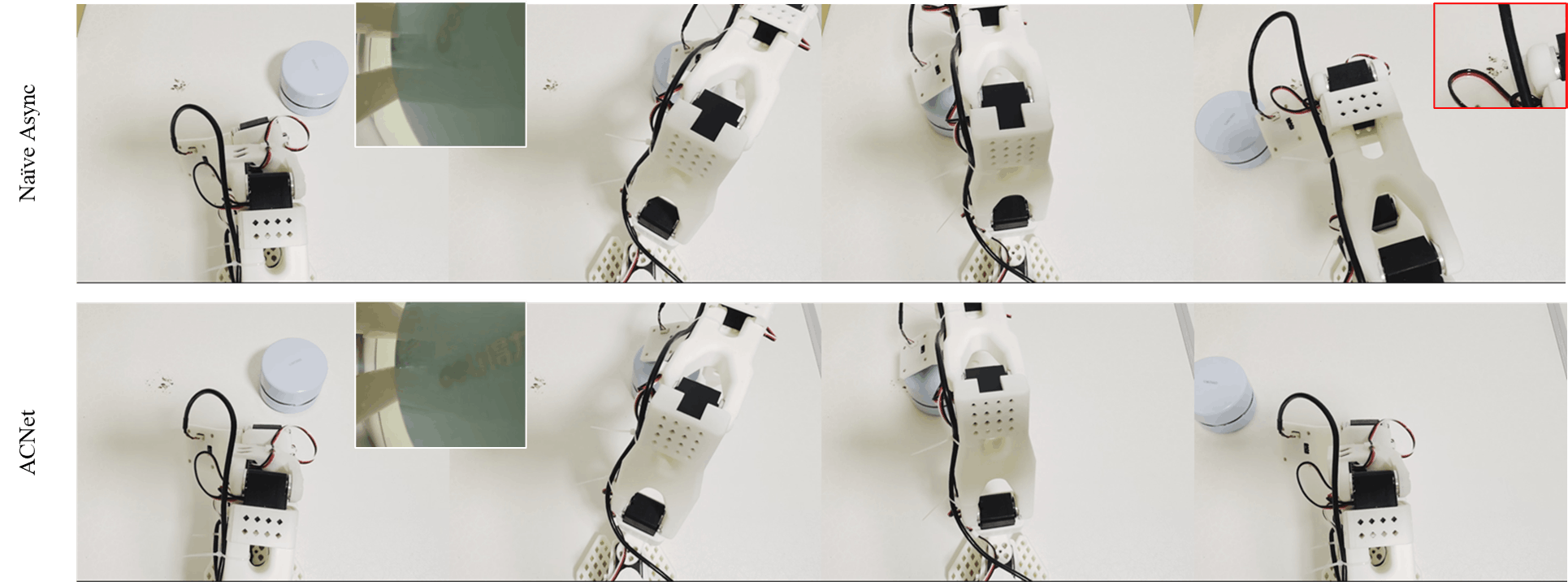}
\vspace{-0.1in}
\caption{\textbf{Top}: Overview of the asynchronous VLA control setting and
the proposed ACNet; \textbf{Bottom}: Real-world asynchronous rollouts on the
SO-ARM101 platform, for the \textit{clean the table} task. In contact-rich
manipulation, inter-chunk jitter in \naiveasync (\textbf{upper}) can shift the
contact point and lead to task failure, whereas ACNet (\textbf{lower})
maintains smoother contact.}
\label{fig:overview}
\end{figure}

Asynchronous inference allows the controller to avoid idle waiting and maintain a higher
effective action frequency~\cite{b4}. Yet this strategy primarily addresses scheduling
rather than modeling. The next action chunk is still generated from an observation that
is stale when the chunk is finally applied, and direct stitching of consecutive chunks
can break continuity at the handoff boundary. The resulting discontinuities often appear
as abrupt action changes, oscillatory corrections, and trajectory jitter, and in
contact-rich or precision-sensitive manipulation they can accumulate into outright task
failure. Existing remedies alleviate this issue only partially. Runtime inpainting or
interpolation can smooth local discontinuities, but such methods are often heuristic and
architecture-dependent. Auxiliary correction heads can compensate for delay-induced
errors, but many are tightly coupled to specific action representations or require
nontrivial policy modification. Training-time delay simulation and full-policy
adaptation offer a more direct route to robustness, but they increase training cost
substantially and weaken the practical advantage of reusing large pretrained VLA
policies.

The dominant error in asynchronous chunked control is treated here as a local handoff
problem rather than a global loss of task understanding. Under inference delay, the
visual and language inputs can often still specify the task and coarse plan correctly, while
the unreliable component is the continuation from the motion already being executed.
This motivates treating delayed VLA control as a boundary conditioning problem at the chunk
handoff. Following this perspective,
\textbf{Action ControlNet (ACNet)} is introduced as a parameter-efficient
module inspired by ControlNet~\cite{b5}. ACNet encodes the short action suffix executed
during inference latency, referred to as the \emph{delay action}, and injects it into a
largely frozen action head as a residual correction. This provides the action expert
with an explicit boundary condition while preserving the task semantics encoded by the
frozen backbone of the perception-language networks.

The contributions are threefold. (I) delay-induced degradation in asynchronous
chunked VLA control is formulated as a boundary-conditioning problem at chunk handoff,
with the executed motion suffix identified as the key conditioning signal. (II) ACNet
is proposed as a lightweight action-head-level residual adapter that conditions a
largely frozen action head on this suffix, requires only a few
trainable parameters, and can be attached as a plug-and-play module to existing chunked
VLA policies with generative action heads such as diffusion and flow matching. (III)
simulation results show that ACNet achieves performance comparable to full
delay-conditioned retraining on Kinetix and Meta-World MT50, and real-world experiments
show visibly reduced transition stuttering on a SO-ARM101 platform.

\section{Related Work}

\subsection{Vision-Language-Action Models and Generative Policies}
Different from earlier explicit skill modeling and
demonstration-based manipulation methods~\cite{b22,b23},
large-scale pretraining has accelerated generalist robot policies that combine
perception, language, and action. Representative VLA systems include
RT-1~\cite{b18}, RT-2~\cite{b1}, Octo~\cite{b19}, OpenVLA~\cite{b6}, and
$\pi_0$~\cite{b7}, which improve semantic generalization, transferability, and
policy capacity across manipulation tasks. In parallel, action models such as
ACT~\cite{b20}, Diffusion Policy~\cite{b2}, flow matching~\cite{b3}, and
Q-Transformer~\cite{b21} improve chunked or generative control. These models are
expressive, but their inference cost makes high-frequency closed-loop deployment
difficult. ACNet complements this line of work by adapting the action head for
delayed asynchronous execution.

\subsection{Asynchronous Inference and Delay-Aware Robot Control}
Asynchronous inference improves throughput by decoupling computation from execution. In
the simplest setting, as used in SmolVLA~\cite{b14} and denoted as \naiveasync, the
robot executes the current chunk while the policy predicts the next one. This removes
idle waiting, but the next chunk is still generated from stale observations, creating
prediction--execution mismatch at handoff. Prior methods address this mismatch through
action inpainting (RTC~\cite{b4}), lightweight correction heads~\cite{b8}, future-state
prediction (VLASH~\cite{b9}), or training-time delay simulation~\cite{b10}. These
methods show that asynchronous control requires temporal alignment inside the policy.
ACNet targets this gap with parameter-efficient residual conditioning in a mostly frozen
action head.

\begin{figure}[t!]
\raggedright
\includegraphics[width=\columnwidth]{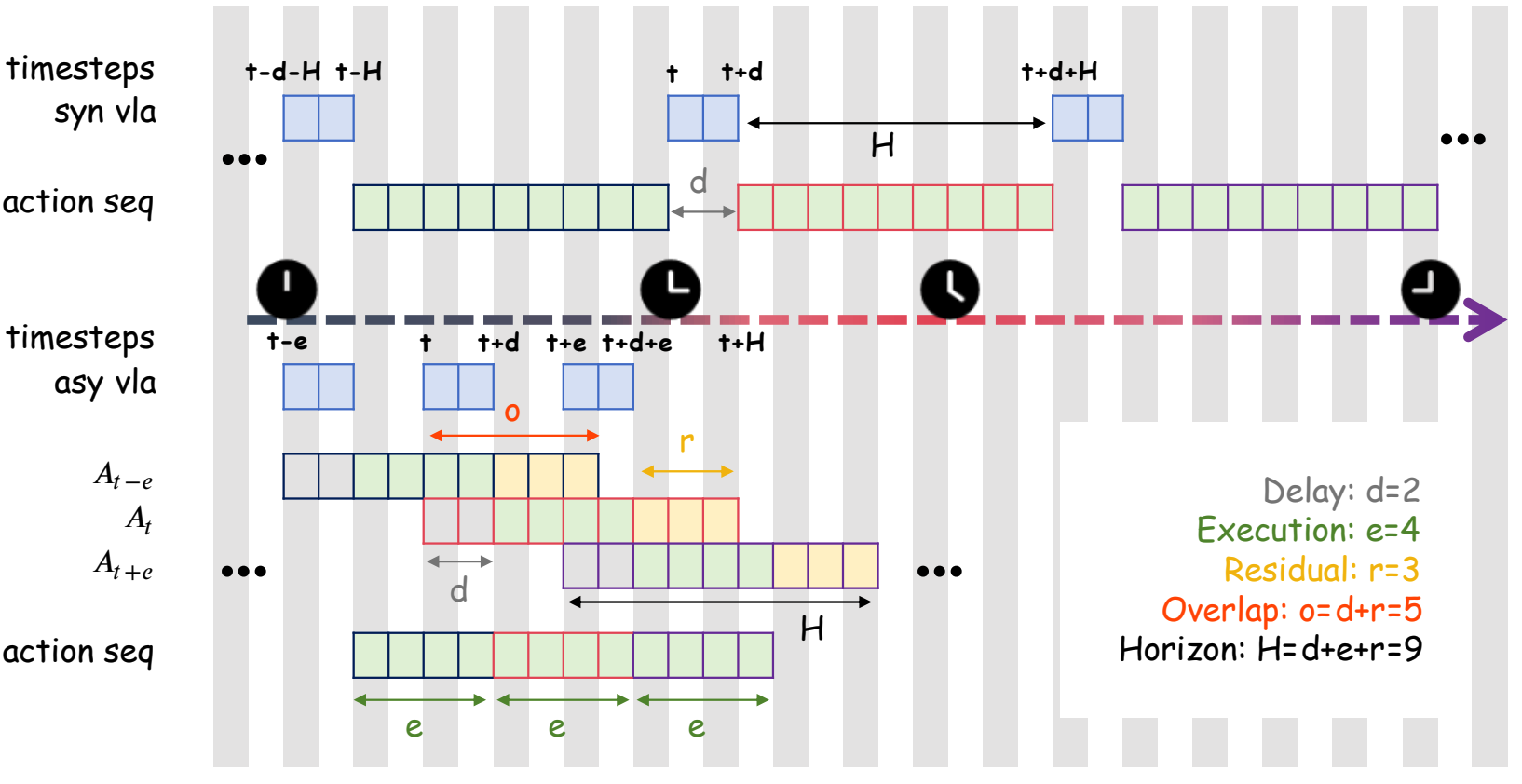}
\vspace{-0.3in}
\caption{Comparisons between synchronous (\textbf{Top})
  and asynchronous (\textbf{Bottom}) execution for chunked VLA policies.}
\label{fig:async_vla}
\end{figure}


\section{Problem Formulation}

Consider a pretrained chunked vision-language-action (VLA) policy $\mathcal{M}_\theta$
with parameters $\theta$. At control step $t$, the policy receives the current
observation $o_t$, which may include visual and proprioceptive inputs, together with a
language instruction $l$, and predicts an action chunk of horizon $H$:
\begin{equation}
    \mathbf{a}_t = \mathcal{M}_\theta(o_t,l)
    = \{a_t^{(0)},a_t^{(1)},\cdots,a_t^{(H-1)}\},
    \quad
    \mathbf{a}_t \in \mathbb{R}^{H\times d_a},
    \label{eq:chunked_vla_policy}
\end{equation}
where $d_a$ is the action dimensionality. The chunk index $t$ denotes the physical
control step at which the corresponding inference is launched.

In asynchronous chunked execution, a new prediction is launched every $e$ control steps,
while the predicted chunk becomes available only after an inference delay of $d$ steps.
During this interval, the robot continues executing the previous chunk. Thus, the chunk
launched at step $t$ is generated from $(o_t,l)$ but is applied only at time $t+d$. Its
first $d$ actions are therefore stale when the chunk becomes available. For boundary
analysis, the chunk horizon is described as:
\begin{equation}
    H = d + e + r,
    \label{eq:horizon_accounting}
\end{equation}
where $e$ is the relaunch interval, $d$ is the inference delay, and $r$ is an optional
future suffix retained beyond the handoff window. This horizon accounting is not
intended as a new execution model; it only makes explicit which part of the previously
predicted chunk has been executed during inference latency. At the handoff time $t+d$,
the first executable action of the new chunk, $a_t^{(d)}$, should continue smoothly from
the last executed action of the previous chunk, $a_{t-e}^{(e+d-1)}$. The relevant motion
context is therefore the executed suffix of the previous chunk:
\begin{equation}
    \mathbf{a}_t^{\mathrm{delay}}
    =
    \{a_{t-e}^{(e)},\cdots,a_{t-e}^{(e+d-1)}\}
    \in \mathbb{R}^{d\times d_a}.
    \label{eq:delay_action}
\end{equation}
This short executed suffix is referred to as the delay action. Since the action head
expects an $H$-step tensor, the delay action is padded to horizon $H$ and denoted by
$\widetilde{\mathbf{a}}_t^{\mathrm{delay}}$.
A pretrained chunked VLA policy can be decomposed into a backbone of perception-language
network followed by an action head as described below:
\begin{equation}
    \mathcal{M}_\theta(o_t,l)
    =
    \mathcal{A}_\psi\!\left(\mathcal{B}_\omega(o_t,l)\right),
    \label{eq:base_vla_decomp}
\end{equation}
where $\theta=\{\omega,\psi\}$, $\mathcal{B}_\omega$ denotes the perception-language
backbone, and $\mathcal{A}_\psi$ denotes the action head. The key observation is that
asynchronous delay primarily introduces a local handoff mismatch rather than a global
failure of task understanding. The observation and language instruction usually still
specify the task semantics and coarse plan, while the unreliable component is the
continuation from the motion already executed during the delay. Therefore, instead of
relearning the full visuomotor mapping, the goal is to construct a delay-aware policy
conditioned on both the task context and the following executed motion context:
\begin{equation}
    \widehat{\mathbf{a}}_t
    =
    \widehat{\mathcal{M}}_{\theta,\eta}
    (o_t,l,\widetilde{\mathbf{a}}_t^{\mathrm{delay}}),
    \label{eq:delay_aware_policy}
\end{equation}
where $\eta$ denotes newly introduced trainable parameters. The adapted policy
$\widehat{\mathcal{M}}_{\theta,\eta}$ is required to preserve the pretrained task semantics
of $\mathcal{M}_\theta$ while reducing the handoff mismatch.

Formally, given a training distribution $\mathcal{D}$ over $(o_t,l,\mathbf{a}_t^*)$ and
a deployment delay distribution $p(d)$, the goal is to seek a parameter-efficient
adaptation $\eta^\star$ that minimizes the expected delayed-control risk as:
\begin{equation}
    \eta^\star
    =
    \arg\min_{\eta:\,|\eta|\ll|\theta|}
    \mathbb{E}
    \left[
        \mathcal{L}_{\mathrm{pred}}
        (\widehat{\mathbf{a}}_t,\mathbf{a}_t^*)
        +
        \lambda
        \mathcal{L}_{\mathrm{bd}}
        (\widehat{\mathbf{a}}_t,\widetilde{\mathbf{a}}_t^{\mathrm{delay}})
    \right],
    \label{eq:delayed_control_risk}
\end{equation}
where $(o_t,l,\mathbf{a}_t^*)\sim\mathcal{D}$ and $d\sim p(d)$. Here,
$\mathcal{L}_{\mathrm{pred}}$ is the base action-generation objective,
$\mathcal{L}_{\mathrm{bd}}$ is a boundary-consistency term, and $\lambda\ge 0$ balances
task fidelity against handoff smoothness. A natural boundary loss is given by:
\begin{equation}
    \mathcal{L}_{\mathrm{bd}}
    =
    \left\|
        \widehat{a}_t^{(d)} - a_{t-e}^{(e+d-1)}
    \right\|_2^2,
    \label{eq:boundary_consistency_loss}
\end{equation}
where $\widehat{a}_t^{(d)}$ is the first executable action of the predicted chunk and
$a_{t-e}^{(e+d-1)}$ is the last executed action of the previous chunk.
Thus, the problem is to adapt a pretrained chunked VLA policy to asynchronous deployment
by using the executed delay action as a boundary condition, while introducing only a
small number of trainable parameters and avoiding full backbone retraining.

\section{Action ControlNet}

\begin{figure}[t]
\centering
\includegraphics[width=\columnwidth]{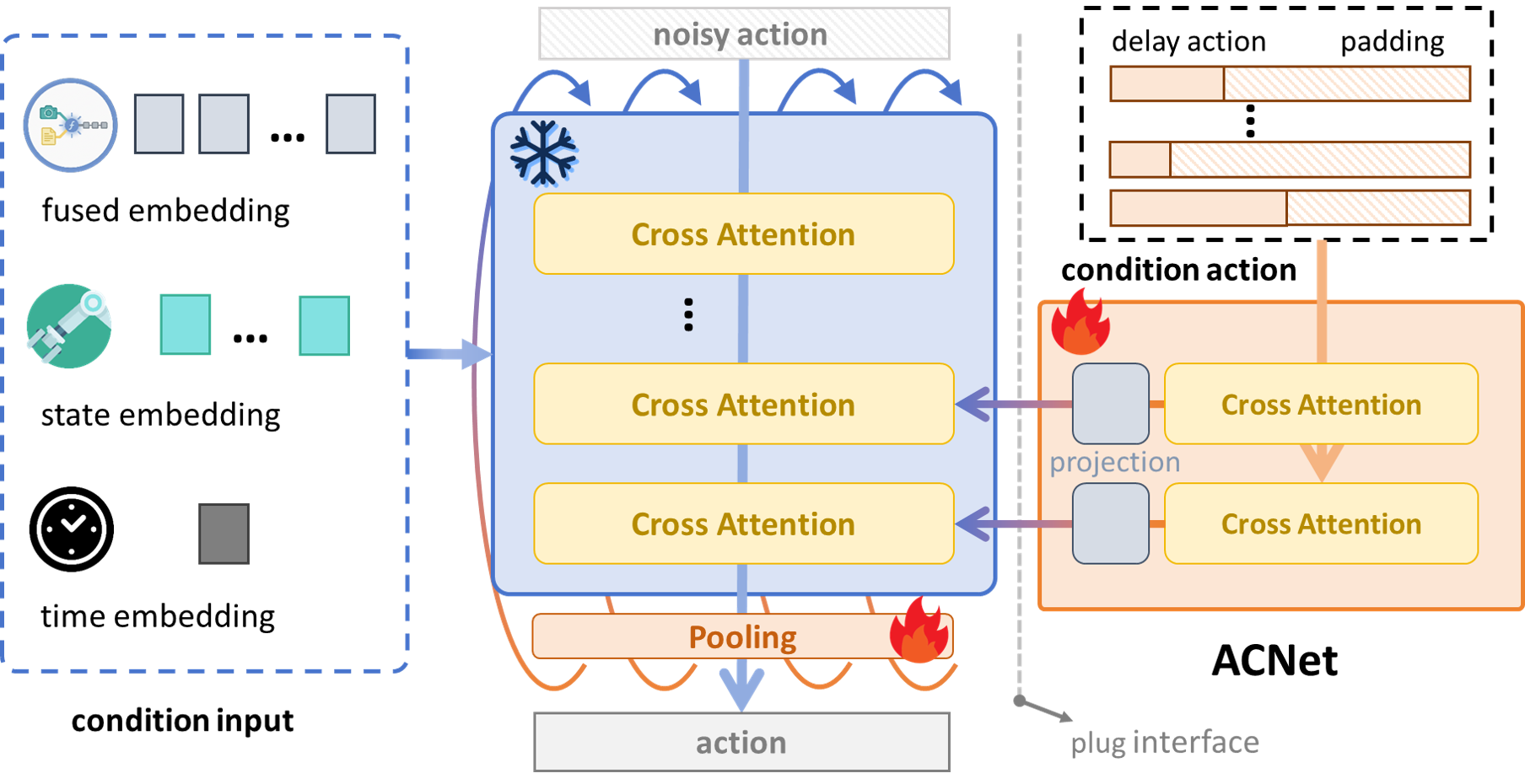}
\vspace{-0.1in}
\caption{Architecture of ACNet. The pretrained perception-language backbone and
main action expert are kept mostly frozen, while the delay action is encoded by
a lightweight side branch and injected into the action head through projection
layers as residual conditioning.}
\label{fig:acnet_model}
\end{figure}

ACNet implements the boundary-conditioned objective by augmenting a pretrained VLA model
only at the action head. This design follows four requirements: preserve the pretrained
backbone (\textbf{R1}), encode the executed suffix (\textbf{R2}), apply the correction
locally in the action head (\textbf{R3}), and remain efficient across sampled delays
(\textbf{R4}). Specifically, the delay-aware policy is factorized as:
\begin{equation}
    \widehat{\mathcal{M}}_{\theta,\eta}
    (o_t,l,\widetilde{\mathbf{a}}_t^{\mathrm{delay}})
    =
    \widehat{\mathcal{A}}_{\psi,\eta}
    \!\left(
    \mathcal{B}_\omega(o_t,l),\widetilde{\mathbf{a}}_t^{\mathrm{delay}}
    \right),
    \label{eq:acnet_policy_factorization}
\end{equation}
where $\mathcal{B}_\omega$ is the frozen perception-language backbone,
$\widehat{\mathcal{A}}_{\psi,\eta}$ is the delay-aware action head, and $\eta$ collects the
new trainable parameters. The parameter set $\eta$ is instantiated as
\textbf{Action ControlNet (ACNet)}, a lightweight residual conditioning branch that
preserves the backbone and adapts only the action head. The overall architecture is
shown in Fig.~\ref{fig:acnet_model}.

\subsection{Delay-Action Encoding}
At time $t+d$, the latent $\mathcal{B}_\omega(o_t,l)$ is available, but the new chunk
must continue the motion already executed during the delay. The observed delay segment
has length $d$, while the policy expects an $H$-step tensor. The executed suffix is
therefore retained explicitly, and the unobserved future positions are padded with
learnable tokens, i.e.,
\begin{equation}
    \widetilde{\mathbf{a}}_t^{\mathrm{delay}}
    =
    \left[
    a_{t-e}^{(e)},\cdots,a_{t-e}^{(e+d-1)},
    \mathbf{p}_{d},\cdots,\mathbf{p}_{H-1}
    \right],
    \label{eq:acnet_padded_delay}
\end{equation}
where $\mathbf{p}_j \in \mathbb{R}^{d_a}$ denotes the padding token at future position
$j$. Learnable padding explicitly marks unavailable slots, avoiding the ambiguity of
zero actions and the artificial variation of noise padding.

ACNet then applies a lightweight transformer encoder $\mathcal{E}_\phi$ to
$\widetilde{\mathbf{a}}_t^{\mathrm{delay}}$, followed by the same terminal temporal pooling
operator used in the action expert, yielding
\begin{equation}
    \mathbf{c}_t
    =
    \mathrm{Pool}\!\left(
    \mathcal{E}_\phi(\widetilde{\mathbf{a}}_t^{\mathrm{delay}})
    \right)
    \in \mathbb{R}^{d_c},
    \label{eq:acnet_condition}
\end{equation}
where $\mathbf{c}_t$ represents the short-horizon executed suffix. Reusing the expert's
terminal pooling aligns this context with the action head's temporal abstraction instead
of introducing a separate control stream.

\begin{figure*}[t!]
\centering
\begin{minipage}[t]{0.49\textwidth}
\centering
\includegraphics[width=\linewidth]{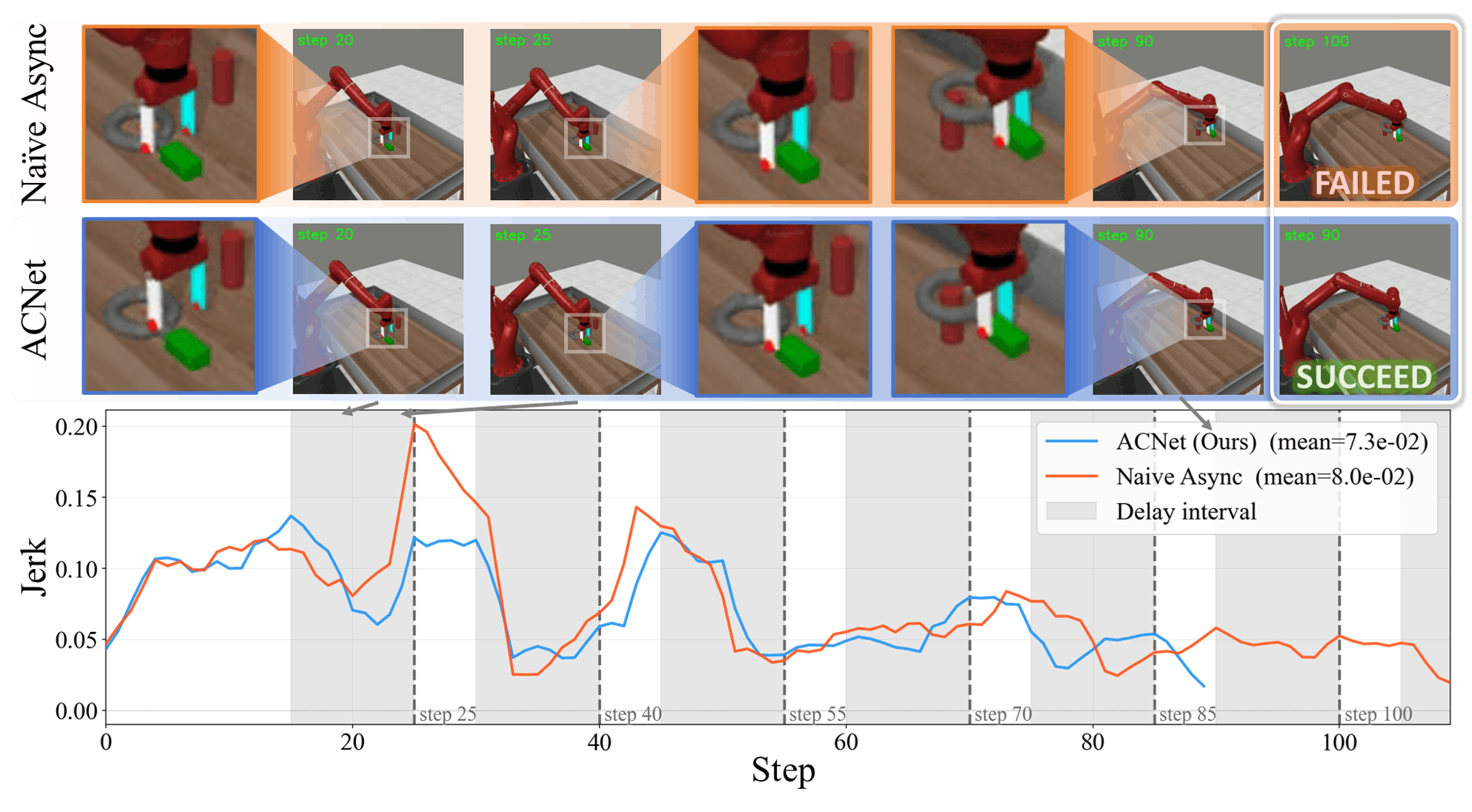}
\end{minipage}\hfill
\begin{minipage}[t]{0.49\textwidth}
\centering
\includegraphics[width=\linewidth]{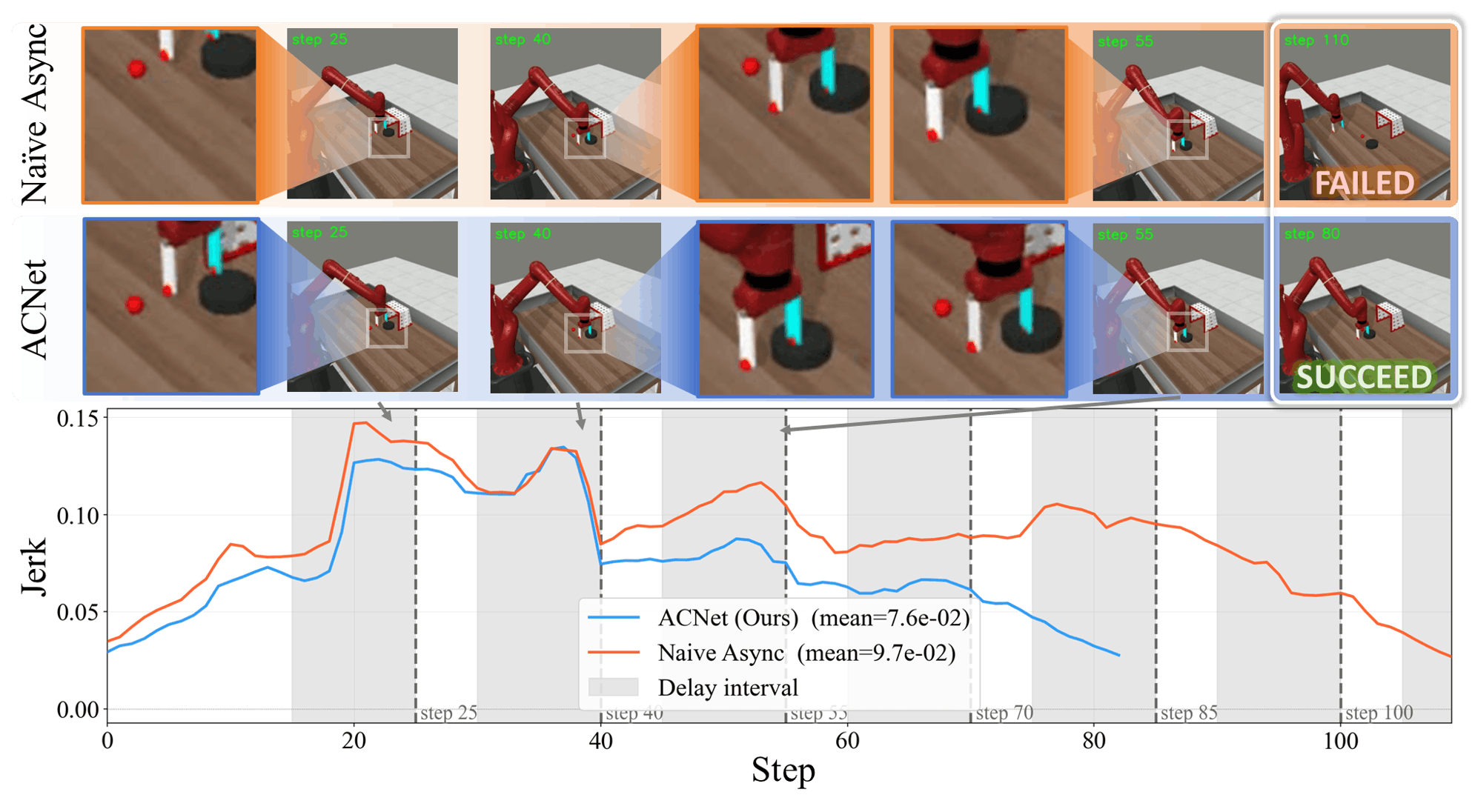}
\end{minipage}
\vspace{-0.1in}
\caption{\textbf{Left}: nut-assembly-v3 task;
  \textbf{Right}: plate-slide-back-v3 task.
Action jerk over the first 50 executed steps for representative
asynchronous rollouts with $H=50$ and $d=10$. Gray regions denote delayed
intervals, and dashed lines denote chunk replacement events. ACNet yields a
flatter jerk profile than \naiveasync around the handoff boundary, indicating
smoother cross-chunk transitions.}
\label{fig:jerk_compare}
\end{figure*}

\subsection{Residual Injection as Local Boundary Correction}
Because the pretrained backbone already encodes the task context, ACNet injects the
delay context only into the action head. The injection is residual so that the original
predictor remains the default mode.
Let $\mathbf{h}_l$ denote the hidden state entering block $l$ of an $L$-layer action
expert, and let $\mathcal{S}\subseteq\{1,\cdots,L\}$ be the set of injected blocks. For
each $l\in\mathcal{S}$, the proposed ACNet applies the following update:
\begin{equation}
    \mathbf{h}_l'
    =
    \mathbf{h}_l + \mathcal{Z}_{\phi,l}(\mathbf{c}_t),
    \label{eq:acnet_residual_injection}
\end{equation}
where $\mathcal{Z}_{\phi,l}$ projects $\mathbf{c}_t$ to the hidden width of the action
expert. This compact side branch can collapse toward zero when the delay cue is
uninformative and acts near the stage where the output trajectory is formed.

Moreover, let $g_l$ denote the downstream mapping from the hidden state in block $l$ to the first
executable action $\widehat{a}_t^{(d)}$. If $g_l$ is differentiable at $\mathbf{h}_l$, then
for a small residual $\mathbf{u}_l=\mathcal{Z}_{\phi,l}(\mathbf{c}_t)$, it holds that:
\begin{equation}
    \widehat{a}_t^{(d)}
    =
    g_l(\mathbf{h}_l+\mathbf{u}_l)
    \approx
    g_l(\mathbf{h}_l) + J_l \mathbf{u}_l,
    \label{eq:first_order_boundary}
\end{equation}
where $J_l$ is the Jacobian of $g_l$ at $\mathbf{h}_l$.
Equation~\eqref{eq:first_order_boundary} provides a local intuition: to first order,
ACNet only needs to produce a residual direction whose downstream effect matches the
desired boundary correction, rather than reconstructing the entire chunk from scratch.
This supports treating asynchronous delay as local boundary conditioning rather than
full visuomotor remapping. In the 8-layer DiT-style action expert of Evo-1~\cite{b12},
this design is instantiated by injecting only the final block, i.e.,
$\mathcal{S}=\{L\}$. Later layers are closest to the action output and therefore most
directly influence the handoff boundary; Sec.~\ref{sec:ablation} validates this choice
empirically.

\subsection{Training Objective}
Because ACNet is attached only to the action head, delay robustness can be learned
without repeatedly forwarding the full backbone for every sampled delay. For a fixed
observation-instruction pair, the visual-language latent produced by
$\mathcal{B}_\omega(o_t,l)$ is invariant to the sampled delay. This latent is therefore
cached once and reused across multiple delay conditions. For a sampled delay $d$, the
corresponding condition vector is given by:
\begin{equation}
    \mathbf{c}_{t,d}
    =
    \mathrm{Pool}\!\left(
    \mathcal{E}_\phi(\widetilde{\mathbf{a}}_{t,d}^{\mathrm{delay}})
    \right),
    \label{eq:acnet_condition_delay}
\end{equation}
where $\widetilde{\mathbf{a}}_{t,d}^{\mathrm{delay}}$ is the padded suffix. This strategy
increases delay coverage while keeping the dominant visual-language computation
amortized.
Furthermore, for the flow-matching action expert in Evo-1, the residual pathway makes the velocity
predictor delay conditioned:
\begin{equation}
    \widehat{\mathbf{v}}_{t,\tau}
    =
    v_{\psi,\eta}^{\mathrm{ACNet}}
    (\mathbf{x}_\tau,\tau,\mathbf{c}_{t,d}),
    \label{eq:acnet_velocity_predictor}
\end{equation}
where $\mathbf{x}_\tau$ is the interpolated action state at flow time $\tau>0$. Relative
to the original flow-matching head, the structural change is the additional residual
conditioning pathway.
It is worth noting that the pretrained backbone and the main action generator remain
unchanged.

Moreover, for the flow-matching training, samples are drawn as $\tau\sim\mathrm{Beta}(2,2)$ and
$\mathbf{z}\sim\mathcal{U}([-1,1]^{H\times d_a})$, define
\begin{equation}
    \mathbf{x}_\tau
    =
    (1-\tau)\mathbf{z}+\tau\mathbf{x}_0,
    \label{eq:flow_interpolation}
\end{equation}
where $\mathbf{x}_0=\mathbf{a}_t^*$ is the target chunk. The training objective is
\begin{equation}
    \mathcal{L}_{\text{FM}}
    =
    \mathbb{E}_{d,\tau,\mathbf{x}_0,\mathbf{z}}
    \left[
    \left\|
    (\mathbf{x}_0-\mathbf{z}) -
    v_{\psi,\eta}^{\mathrm{ACNet}}(\mathbf{x}_\tau,\tau,\mathbf{c}_{t,d})
    \right\|_2^2
    \right],
    \label{eq:acnet_training_loss}
\end{equation}
where the expectation is taken over sampled delays, flow times, target chunks, and
initial action samples. This objective instantiates the delayed-control risk in
\eqref{eq:delayed_control_risk} with a flow-matching prediction loss, while ACNet
enforces boundary-aware conditioning through the residual pathway in
\eqref{eq:acnet_residual_injection}.

\section{Experiments}

\subsection{Evaluation Protocol}
ACNet is evaluated on two asynchronous manipulation benchmarks: the Kinetix setup from
RTC~\cite{b4} in the Kinetix simulator~\cite{b15}, and the full Meta-World MT50
suite~\cite{b13}. All experiments compare four deployment schemes under the same
asynchronous execution protocol:
\textbf{\naiveasync}, which directly stitches consecutive chunks;
\textbf{RTC}~\cite{b4}, which freezes committed actions and inpaints the
remaining suffix; \textbf{Training-RTC}~\cite{b10}, which performs delay-conditioned
training; and \textbf{ACNet}, the proposed delay-aware adapter.

On Kinetix, the RTC evaluation protocol is followed~\cite{b4}. The RTC and Training-RTC
baselines are based on a $\pi_0$ policy~\cite{b7}: to keep training exposure comparable,
\naiveasync and RTC share the same backbone trained for 32 epochs, while Training-RTC
follows the schedule in~\cite{b10}, with 24 epochs of standard training and 8 additional
epochs of delay-conditioned adaptation. For ACNet, the Evo-1 backbone is
used~\cite{b12}: stage 1 is trained for 1 epoch and stage 2 for 24 epochs, after which
the backbone is frozen and ACNet together with the terminal pooling layer is trained for
8 epochs. Delays $d \in \{0,1,2,3,4\}$ are evaluated.

On Meta-World, all 50 MT50 tasks are evaluated. The RTC and Training-RTC baselines use
their $\pi_0$-based implementations, whereas ACNet uses Evo-1~\cite{b12} as the
backbone. For the benchmark-wide results in Table~\ref{tab:metaworld_main}, the
prediction horizon is $H=50$, the execution interval is $e=25$, and the evaluated delays
are $d \in \{0,5,10,15\}$. The representative jerk plots in Fig.~\ref{fig:jerk_compare}
visualize rollouts on
\textit{nut-assembly-v3} and \textit{plate-slide-back-v3} with $H=50$ and
$d=10$. All Meta-World experiments are run on a single RTX 4080 SUPER.

\subsection{Metrics}
Task success rate is the primary metric on both benchmarks. To expose the
efficiency--robustness trade-off more clearly, trainable parameters are also reported on
Kinetix, measured as the percentage of total model parameters updated during adaptation.
Frozen parameters are not counted as trainable. End-to-end inference latency and
achieved control frequency are also reported on Meta-World.
For representative Meta-World rollouts, translational jerk is also reported. For
$\mathbf{a}_t = (a_t^x,a_t^y,a_t^z)$, the step-wise jerk of the resulting trajectory
is defined as follows:
\begin{equation}
    J_t = \frac{1}{3}\sum_{q\in\{x,y,z\}}
    \left|a_{t+2}^{q} - 2a_{t+1}^{q} + a_t^{q}\right|,
\end{equation}
with lower values indicating smoother action evolution.

\begin{table}[!tbp]
\caption{Kinetix results under asynchronous inference delay.}
\label{tab:kinetix_main}
\centering
\renewcommand{\arraystretch}{1.15}
\setlength{\tabcolsep}{4pt}
\resizebox{\columnwidth}{!}{%
\begin{tabular}{lccccccc}
\toprule
\multirow{2}{*}{\textbf{Method}} &
\multicolumn{5}{c}{\textbf{Success rate under delay $d$}} &
\multirow{2}{*}{\textbf{Avg. $(d>0)$}} &
\multirow{2}{*}{\textbf{Params. (\%)}} \\
\cmidrule(lr){2-6}
& \textbf{0} & \textbf{1} & \textbf{2} & \textbf{3} & \textbf{4} & & \\
\midrule
\naiveasync     & 0.89 & 0.74 & 0.69 & 0.55 & 0.46 & 0.61 & \textbf{0} \\
RTC             & \textbf{0.91} & 0.75 & 0.80 & 0.72 & 0.61 & 0.72 & \textbf{0} \\
Training-RTC    & 0.89 & \textbf{0.88} & \underline{0.83} & \textbf{0.79} & \textbf{0.70} & \textbf{0.80} & 100 \\
ACNet    & \underline{0.90} & \underline{0.87} & \textbf{0.84} & \underline{0.76} & \underline{0.68} & \underline{0.79} & $\sim$20 \\
\bottomrule
\end{tabular}%
}
\end{table}

\begin{table}[!tbp]
\caption{Meta-World MT50 results under asynchronous delay.}
\label{tab:metaworld_main}
\centering
\footnotesize
\renewcommand{\arraystretch}{1.15}
\setlength{\tabcolsep}{7pt}
\resizebox{\columnwidth}{!}{%
\begin{tabular}{lccccccc}
\toprule
\multirow{2}{*}{\textbf{Method}} &
\multicolumn{4}{c}{\textbf{Success rate under delay $d$}} &
\multirow{2}{*}{\textbf{Avg.}} &
\multirow{2}{*}{\textbf{Lat.}} &
\multirow{2}{*}{\textbf{Freq.}} \\
\cmidrule(lr){2-5}
& \textbf{0} & \textbf{5} & \textbf{10} & \textbf{15} & & & \\
\midrule
\naiveasync
& \underline{0.80} & 0.71 & 0.70 & 0.70  & 0.70
& \textbf{73 ms} & \textbf{13.6 Hz} \\
RTC
& 0.79 & 0.72 & 0.72 & 0.71 & 0.71
& 159 ms & 6.28 Hz \\
Training-RTC
& \underline{0.80} & \textbf{0.77} & \textbf{0.74} & \textbf{0.73}
& \textbf{0.74}
& 134 ms & 7.46 Hz \\
ACNet
& \textbf{0.81} & \underline{0.76} & \textbf{0.74}
& \underline{0.73} & \underline{0.74}
& \underline{91 ms} & \underline{11.0 Hz} \\
\bottomrule
\end{tabular}
}
\end{table}

\subsection{Quantitative Results}
Table~\ref{tab:kinetix_main} reports Kinetix results. Averaged over delayed settings,
ACNet reaches 0.79 success, compared with 0.72 for RTC and 0.61 for
\naiveasync, while remaining close to Training-RTC at 0.80. This robustness is
obtained while training only around $20\%$ of all model parameters, whereas Training-RTC
updates 100\%. Thus, although the average success gap between ACNet and Training-RTC is
small, ACNet achieves comparable robustness with an around $80\%$ lower
trainable-parameter fraction. In Table~\ref{tab:kinetix_main}, Params. (\%) denotes the
percentage of total model parameters updated during adaptation.
Table~\ref{tab:metaworld_main} shows the same trend on Meta-World MT50. ACNet achieves
0.74 average success, exceeding \naiveasync and RTC and matching Training-RTC, while
reducing latency to 91\,ms versus 159\,ms for RTC and 134\,ms for Training-RTC. Its
achieved control frequency is 11.0\,Hz, compared with 6.28\,Hz and 7.46\,Hz for those
two baselines. This latency advantage is mainly due to the underlying model choice
rather than the ACNet side branch being intrinsically faster than delay-conditioned
training: RTC and Training-RTC are based on $\pi_0$, whereas ACNet is built on the
lightweight Evo-1 backbone. On the same backbone, the proposed ACNet would add
a small encoder and projection overhead to a standard forward pass.
As shown in the above results, this
overhead is outweighed by the lower end-to-end runtime of Evo-1.

\begin{table}[!tbp]
\caption{Performance on SO-ARM101 over 10 trials per task.}
\label{tab:real_world_success}
\centering
\renewcommand{\arraystretch}{1.15}
\setlength{\tabcolsep}{5pt}
\resizebox{\columnwidth}{!}{%
\begin{tabular}{lccc}
\toprule
\textbf{Method} &
\textbf{Put cube into box} &
\textbf{Clean table} &
\textbf{Overall} \\
\midrule
\naiveasync   & 9/10 (90\%)  & 8/10 (80\%)  & 17/20 (85\%) \\
ACNet & 10/10 (100\%) & 10/10 (100\%) & 20/20 (100\%) \\
\bottomrule
\end{tabular}%
}
\end{table}

\begin{figure}[!tbp]
\centering
\includegraphics[width=0.98\columnwidth]{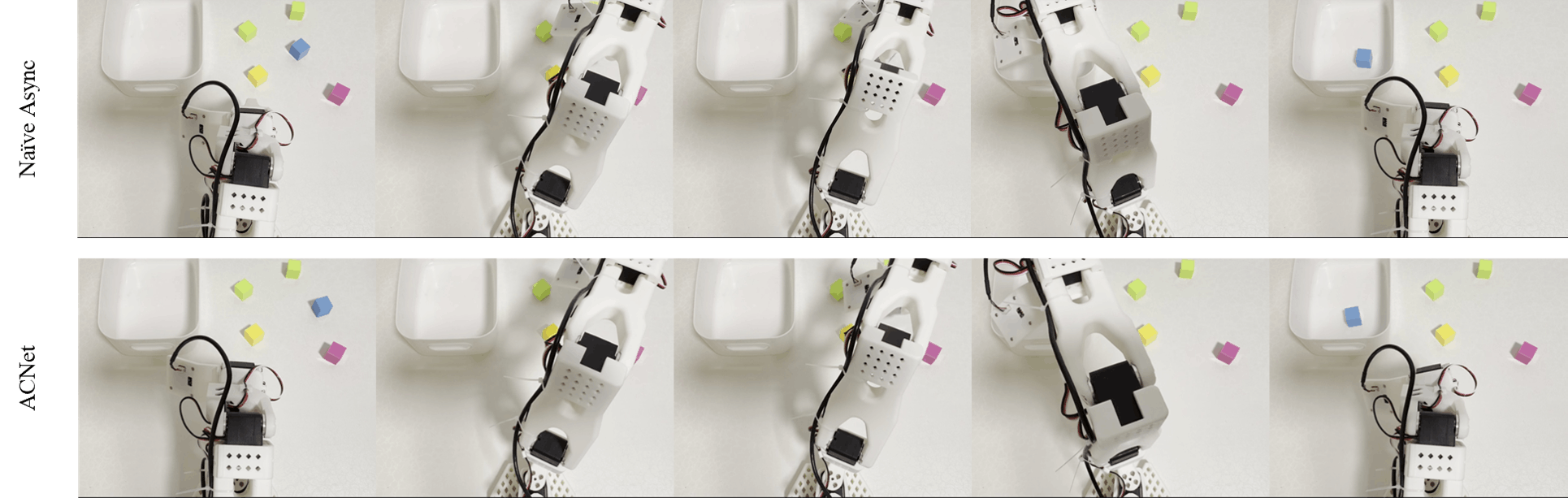}
\vspace{-0.1in}
\caption{Representative real-world asynchronous rollouts on the SO-ARM101
platform, for the \textit{put the blue cube into the box}. Snapshots
are ordered from left to right. The upper sequence shows \naiveasync and the
lower sequence shows ACNet.}
\label{fig:real_tasks}
\end{figure}

\subsection{Real-World Experiments}
\label{sec:real_world}
ACNet is further evaluated on a real tabletop manipulation platform built on the
SO-ARM101 robot arm with a target box, colored cubes, and a tabletop vacuum. The
real-world dataset contains 50 training rollouts. The model is optimized for 10 epochs
and evaluated on \textit{put the blue cube into the box} and
\textit{clean the table} over 10 trials per task under the same asynchronous
protocol as simulation.
The quantitative results are summarized in Table~\ref{tab:real_world_success}. ACNet
achieves 20/20 total successes across the two tasks, whereas \naiveasync achieves 17/20.
The qualitative rollouts in Fig.~\ref{fig:real_tasks} show the same pattern: \naiveasync
has larger oscillations near chunk transitions, while ACNet maintains smoother transfer
and more stable contact during cleaning.

\subsection{Trajectory Smoothness Analysis}
\label{sec:traj_smoothness}
To examine motion quality beyond success rate, representative rollouts from
\textit{nut-assembly-v3} and \textit{plate-slide-back-v3} with $H=50$ and $d=10$ are
analyzed. The jerk traces in Fig.~\ref{fig:jerk_compare} show that ACNet maintains a
flatter profile around delayed intervals and chunk replacement events. In both tasks,
\naiveasync exhibits stronger transition-induced jitter, whereas ACNet produces smoother
cross-chunk continuation, consistent with the success gains in
Table~\ref{tab:metaworld_main}.

\begin{figure}[!tbp]
\centering
\includegraphics[width=\columnwidth]{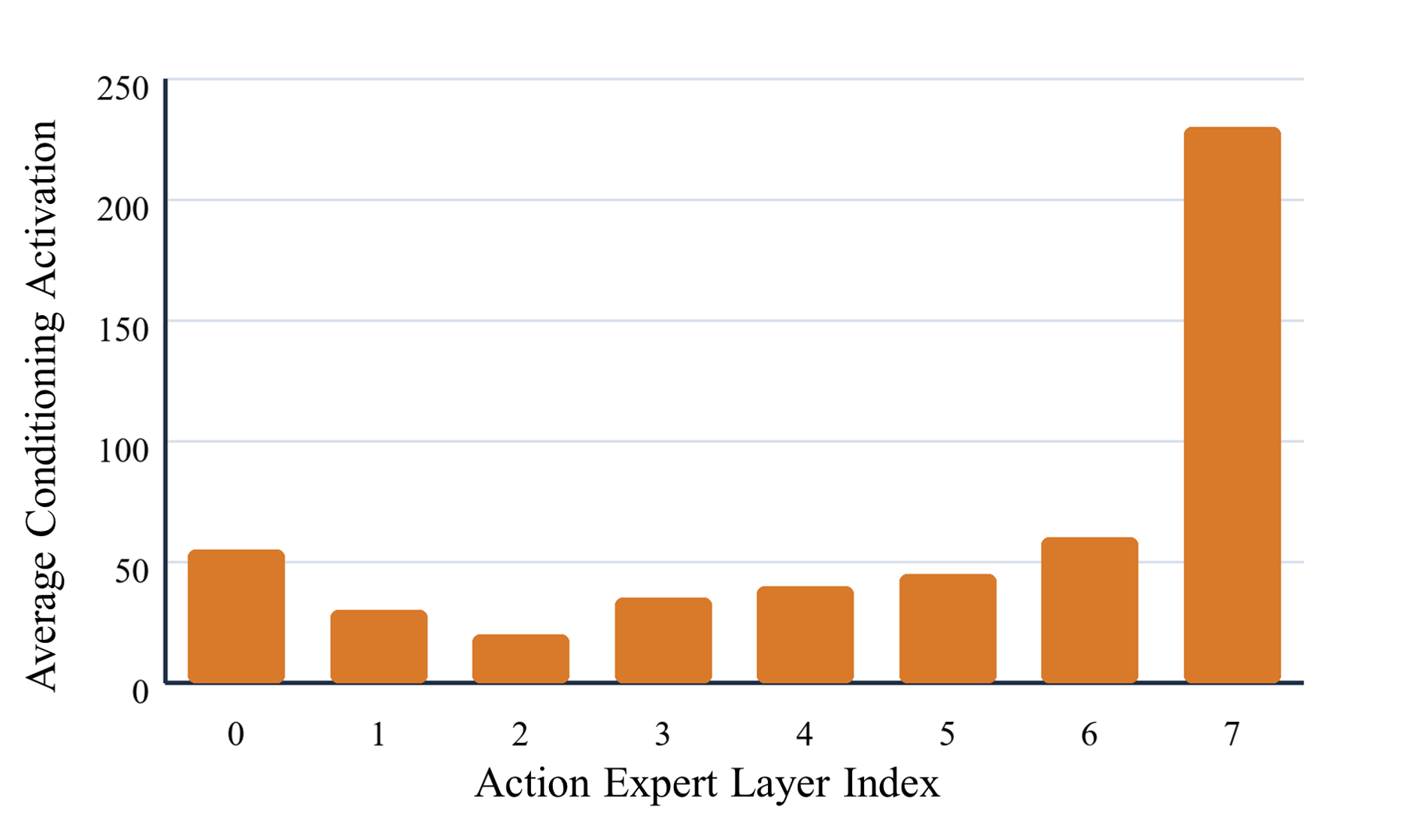}
\vspace{-0.3in}
\caption{Average layer-update magnitude across the 8-layer Evo-1 action
expert over $K=1000$ flow-matching integration steps.}
\label{fig:layer_sensitivity}
\end{figure}

\begin{figure}[!tbp]
\centering
\includegraphics[width=\columnwidth]{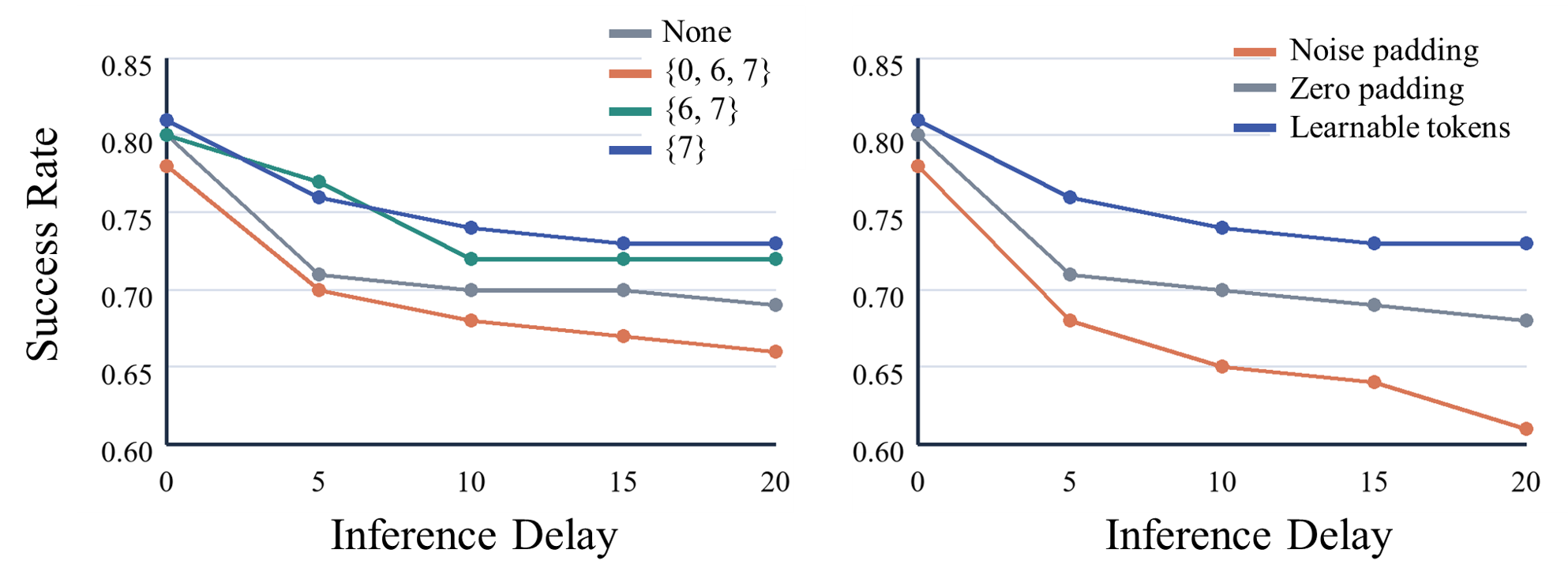}
\vspace{-0.3in}
\caption{Ablations on ACNet design choices in Meta-World MT50 with the Evo-1
backbone. \textbf{Left}: Conditioning only the final action-expert block yields the best
robustness across delays.
\textbf{Right}: Learnable tokens consistently outperform noise
padding and zero padding.}
\label{fig:ablation_panels}
\end{figure}

\subsection{Ablation Study}
\label{sec:ablation}
The ablation study examines which design choices are responsible for the performance
gains of ACNet. All ablations are conducted on Meta-World MT50 using the Evo-1 backbone.
Two questions are considered: \emph{where} the delay-conditioned residual should be
injected, and \emph{how} the unobserved future portion of the delay action should be
represented.
To determine where delay-aware conditioning is most effective, the mean token-update
magnitude of each transformer block is recorded over $K=1000$ flow-matching steps.
Fig.~\ref{fig:layer_sensitivity} shows that the final block has the largest activity,
indicating that late layers contribute most to the final refinement of the predicted
chunk.
The injection ablation in Fig.~\ref{fig:ablation_panels}(a) confirms this choice:
final-block conditioning is best across delays, while early-layer injection, especially
at layer 0, degrades robustness. This supports using the delay action as a local
boundary cue rather than an early perturbation to the task representation.
Fig.~\ref{fig:ablation_panels}(b) shows that learnable padding tokens outperform zero
and random-noise padding at every tested delay. The result supports the
minimal-intervention design of ACNet: represent the delay action unambiguously and
inject it as a late residual correction close to the action output.

\section{Conclusion}
This paper presented Action ControlNet (ACNet), a lightweight delay-aware adapter for
asynchronous deployment of chunked vision-language-action policies. By conditioning a
largely frozen action head on the executed motion suffix, ACNet reduces handoff mismatch
under inference delay while preserving the pretrained perception-language backbone.
Experiments on Kinetix, Meta-World MT50, and a real SO-ARM101 platform show that ACNet
improves delayed-control robustness and trajectory smoothness with substantially lower
adaptation cost than full delay-conditioned retraining. Future work will extend this
boundary-conditioning framework to richer delay signals, larger and time-varying delays,
and broader classes of real-time control policies.

\bibliographystyle{IEEEtran}
\bibliography{references}

@inproceedings{b1,
  author = {B. Zitkovich and others},
  title = {{RT-2: Vision-Language-Action Models Transfer Web Knowledge to Robotic Control}},
  booktitle = {Proc. Conf. Robot Learning (CoRL)},
  pages = {2165--2183},
  year = {2023}
}

@inproceedings{b2,
  author = {C. Chi and others},
  title = {{Diffusion Policy: Visuomotor Policy Learning via Action Diffusion}},
  booktitle = {Proc. Robotics: Science and Systems (RSS)},
  year = {2023}
}

@inproceedings{b3,
  author = {Y. Lipman and others},
  title = {{Flow Matching for Generative Modeling}},
  booktitle = {Proc. Int. Conf. Learning Representations (ICLR)},
  year = {2023}
}

@inproceedings{b4,
  author = {K. Black and M. Y. Galliker and S. Levine},
  title = {{Real-Time Execution of Action Chunking Flow Policies}},
  booktitle = {Proc. Adv. Neural Inf. Process. Syst. (NeurIPS)},
  year = {2025}
}

@inproceedings{b5,
  author = {L. Zhang and A. Rao and M. Agrawala},
  title = {{Adding Conditional Control to Text-to-Image Diffusion Models}},
  booktitle = {Proc. IEEE/CVF Int. Conf. Computer Vision (ICCV)},
  pages = {3836--3847},
  year = {2023}
}

@inproceedings{b6,
  author = {M. Kim and others},
  title = {{OpenVLA: An Open-Source Vision-Language-Action Model}},
  booktitle = {Proc. Conf. Robot Learning (CoRL)},
  pages = {2679--2713},
  year = {2025}
}

@inproceedings{b7,
  author = {K. Black and others},
  title = {{{$\pi_0$}: A Vision-Language-Action Flow Model for General Robot Control}},
  booktitle = {Proc. Robotics: Science and Systems (RSS)},
  year = {2025}
}

@misc{b8,
  author = {K. Sendai and M. Alvarez and T. Matsushima and Y. Matsuo and Y. Iwasawa},
  title = {{Leave No Observation Behind: Real-Time Correction for VLA Action Chunks}},
  note = {arXiv preprint arXiv:2509.23224},
  year = {2025}
}

@misc{b9,
  author = {J. Tang and Y. Sun and Y. Zhao and S. Yang and Y. Lin and Z. Zhang and J. Hou and Y. Lu and Z. Liu and S. Han},
  title = {{VLASH: Real-Time VLAs via Future-State-Aware Asynchronous Inference}},
  note = {arXiv preprint arXiv:2512.01031},
  year = {2025}
}

@misc{b10,
  author = {K. Black and A. Z. Ren and M. Equi and S. Levine},
  title = {{Training-Time Action Conditioning for Efficient Real-Time Chunking}},
  note = {arXiv preprint arXiv:2512.05964},
  year = {2025}
}

@inproceedings{b12,
  author = {T. Lin and Y. Zhong and Y. Du and J. Zhang and J. Liu and Y. Chen and E. Gu and Z. Liu and H. Cai and Y. Zou and L. Zou and Z. Zhou and G. Li and B. Zhao},
  title = {{Evo-1: Lightweight Vision-Language-Action Model with Preserved Semantic Alignment}},
  booktitle = {Proc. IEEE/CVF Conf. Computer Vision and Pattern Recognition (CVPR)},
  pages = {13397--13406},
  year = {2026}
}

@inproceedings{b13,
  author = {T. Yu and D. Quillen and Z. He and R. Julian and K. Hausman and C. Finn and S. Levine},
  title = {{Meta-World: A Benchmark and Evaluation for Multi-Task and Meta Reinforcement Learning}},
  booktitle = {Proc. Conf. Robot Learning (CoRL)},
  pages = {1094--1100},
  year = {2020}
}

@misc{b14,
  author = {M. Shukor and others},
  title = {{SmolVLA: A Vision-Language-Action Model for Affordable and Efficient Robotics}},
  note = {arXiv preprint arXiv:2506.01844},
  year = {2025}
}

@inproceedings{b15,
  author = {M. Matthews and M. Beukman and C. Lu and J. N. Foerster},
  title = {{Kinetix: Investigating the Training of General Agents through Open-Ended Physics-Based Control Tasks}},
  booktitle = {Proc. Int. Conf. Learning Representations (ICLR)},
  year = {2025}
}

@inproceedings{b18,
  author = {A. Brohan and others},
  title = {{RT-1: Robotics Transformer for Real-World Control at Scale}},
  booktitle = {Proc. Robotics: Science and Systems (RSS)},
  year = {2023}
}

@inproceedings{b19,
  author = {{Octo Model Team} and others},
  title = {{Octo: An Open-Source Generalist Robot Policy}},
  booktitle = {Proc. Robotics: Science and Systems (RSS)},
  year = {2024}
}

@inproceedings{b20,
  author = {T. Z. Zhao and V. Kumar and S. Levine and C. Finn},
  title = {{Learning Fine-Grained Bimanual Manipulation with Low-Cost Hardware}},
  booktitle = {Proc. Robotics: Science and Systems (RSS)},
  year = {2023}
}

@inproceedings{b21,
  author = {Y. Chebotar and others},
  title = {{Q-Transformer: Scalable Offline Reinforcement Learning via Autoregressive Q-Functions}},
  booktitle = {Proc. Conf. Robot Learning (CoRL)},
  pages = {3909--3928},
  year = {2023}
}

@article{b22,
  author = {M. Guo and M. B{\"u}rger},
  title = {{Geometric Task Networks: Learning Efficient and Explainable Skill Coordination for Object Manipulation}},
  journal = {IEEE Transactions on Robotics},
  volume = {38},
  number = {3},
  pages = {1723--1734},
  year = {2022}
}

@inproceedings{b23,
  author = {A. T. Le and M. Guo and N. van Duijkeren and L. Rozo and R. Krug and A. G. Kupcsik and M. B{\"u}rger},
  title = {{Learning Forceful Manipulation Skills from Multi-Modal Human Demonstrations}},
  booktitle = {Proc. IEEE/RSJ Int. Conf. Intelligent Robots and Systems (IROS)},
  pages = {7770--7777},
  year = {2021}
}

\end{document}